\pdfoutput=1

\documentclass[11pt]{article}

\usepackage[]{ACL2023}

\usepackage{times}
\usepackage{latexsym}

\usepackage[T1]{fontenc}

\usepackage[utf8]{inputenc}

\usepackage{microtype}

\usepackage{inconsolata}

\usepackage{times}
\usepackage{latexsym}
\usepackage[utf8]{inputenc}
\usepackage{bm}
\usepackage{graphicx}
\usepackage{amsmath}
\usepackage{amsfonts}
\usepackage{booktabs}
\usepackage{multirow}
\usepackage{enumitem}
\usepackage{subcaption}
\usepackage{flushend}

\newcommand{\wxyedit}[1]{\textcolor{black}{#1}}
\newcommand{\wxytodo}[1]{\textcolor{black}{#1}}
\newcommand{\ModelName}{\texttt{ProCNet}}
\newcommand{\MinName}{HDM}

\title{Document-Level Multi-Event Extraction with Event Proxy Nodes and Hausdorff Distance Minimization}

\author{
    Xinyu Wang\textsuperscript{\rm1,2}, 
    Lin Gui\textsuperscript{\rm2}, 
    Yulan He\textsuperscript{\rm1,2,3} \\
  \textsuperscript{1}Department of Computer Science, University of Warwick \\
  \textsuperscript{2}Department of Informatics, King's College London\\
  \textsuperscript{3}The Alan Turing Institute\\
  \texttt{Xinyu.Wang.11@warwick.ac.uk} \\
  \texttt{\{lin.1.gui, yulan.he\}@kcl.ac.uk} \\
  }

\begin{document}

\maketitle

\begin{abstract}

Document-level multi-event extraction aims to extract the structural information from a given document automatically. 
Most recent approaches usually involve two steps: (1) modeling entity interactions; (2) decoding entity interactions into events.
However, such approaches ignore a global view of inter-dependency of multiple events.
Moreover, an event is decoded by iteratively merging its related entities as arguments, which might suffer from error propagation and is computationally inefficient.
In this paper, we propose an alternative approach for document-level multi-event extraction with event proxy nodes and Hausdorff distance minimization. 
The event proxy nodes, representing pseudo-events, are able to build connections 
with other event proxy nodes, essentially capturing global information. 
The Hausdorff distance makes it possible to compare the similarity between the set of predicted events and the set of ground-truth events.
By directly minimizing Hausdorff distance, the model is trained towards the global optimum directly, which improves performance and reduces training time.
Experimental results show that our model outperforms previous state-of-the-art method 
in F1-score on two datasets with only a fraction of 
training time. \footnote{Code is available at \href{https://github.com/xnyuwg/procnet}{https://github.com/xnyuwg/procnet}}


\end{abstract}

\section{Introduction}

\begin{figure}[t]
	\centering 
	\centerline{\includegraphics[width=0.49\textwidth]{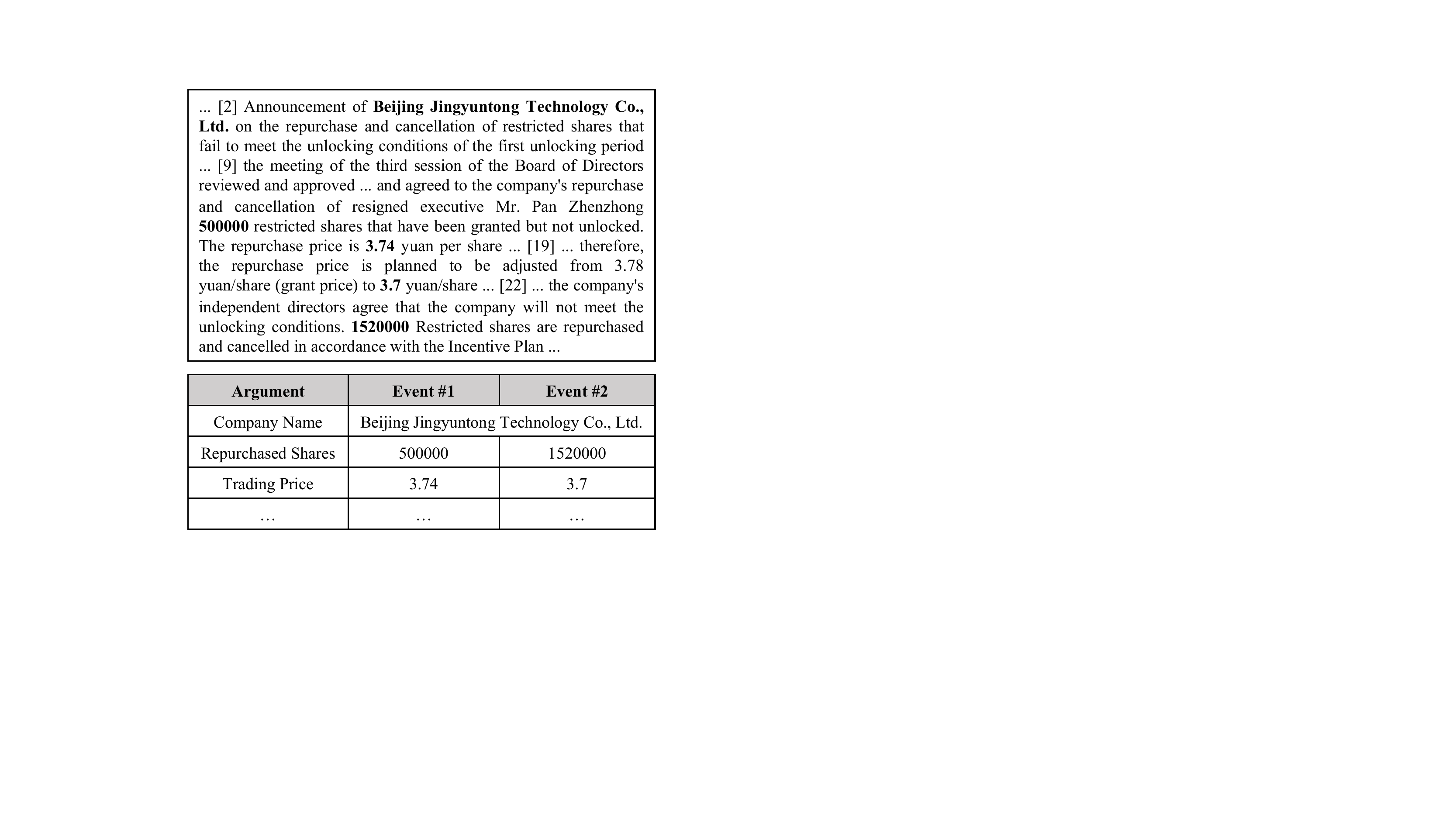}}
	\caption{An example of a document that contains two events. [$\cdot$] denotes the sentence numbering. Words highlighted in colors denote different entities.} 
	\label{fig:example}
\end{figure}

Event extraction aims to identify event triggers with certain types and extract their corresponding arguments from text. Much research has been done on sentence-level event extraction \citep{
du-cardie-2020-event, lin-etal-2020-joint, lu-etal-2021-text2event}.
In recent years, there have been growing interests in tackling the more challenging task of document-level 
multi-event extraction, where an event is represented by a cluster of arguments, which may scatter across multiple sentences in a document.
Also, multiple events in the same document may share some common entities. For example, as shown in Figure~\ref{fig:example}, the two events, \emph{Equity Pledge} and \emph{Equity Freeze}, have their arguments scattered across the document. The same entity mentions, \emph{Yexiang Investment Management Co., Ltd.} and \emph{13.07\%}, are involved in both events, with the former taking different argument roles (`\emph{Pledger}' and `\emph{Equity Holder}'), while the latter having the same argument role (`\emph{Total Holding Ratio}'). In such a setup, an event is not associated with a specific event trigger word or phrase, as opposed to the common setup in sentence-level event extraction. 
These challenges make it difficult to distinguish various events and link entities to event-specific argument roles. 

Document-level multi-event extraction 
can be typically formulated as a table-filling task that fills the correct entities into a pre-defined event schema as shown in Figure~\ref{fig:example}.
Here, an event is essentially represented by a cluster of arguments.
Existing approaches \citep{zheng-etal-2019-doc2edag, yang-etal-2021-document, huang-jia-2021-exploring-sentence, xu-etal-2021-document, liang-etal-2022-raat} usually involve two steps: (1) first model the entity interactions based on contextual representations; 
(2) then design a decoding strategy to 
decode the entity interactions into events and arguments. For example, \citet{zheng-etal-2019-doc2edag} and \citet{xu-etal-2021-document} transformed this task into 
sequential path-expanding sub-tasks. Each sub-task expands a path sequentially by gradually merging entities in a pre-defined order of event argument roles.

The aforementioned approaches suffer from the following limitations:
(1) They decode events from entity information and tend to produce local optimal results without considering the inter-dependency of multiple events globally in a document. 
(2) Event decoding by iteratively merging entities 
suffers from error propagation that an event type or an entity that has been incorrectly classified cannot be corrected later. (3) Every decoding decision requires iterating all entity mentions in a document, which is computationally inefficient.

To address the above limitations, we propose an alternative approach for document-level multi-event extraction with event proxy nodes and Hausdorff distance minimization, named as \textbf{Pro}xy Nodes \textbf{C}lustering \textbf{Net}work (\ModelName). The event proxy nodes aim to capture the global information 
among events in a document. The Hausdorff distance 
makes it possible to optimize the training loss defined as the difference between the generated events and the gold standard event annotations directly. This is more efficient compared to existing 
decoding approaches.


Our method involves two main steps: \emph{Event Representation Learning} and \emph{Hausdorff Distance Minimization}.
For \emph{Event Representation Learning}, we create a number of proxy nodes, each of which represents a pseudo-event, and build a graph to update proxy nodes. 
Entities mentioned in text are treated as nodes connecting to the proxy nodes. 
All the proxy nodes are interconnected to allow information exchange among the potential events. 
We employ a Hypernetwork Graph Neural Network (GNN) \citep{iclr/HaDL17} for updating proxy node representations. 
After \emph{Event Representation Learning}, each proxy node 
essentially resides in a new event-level metric space by aggregating information from the entity-level space.

For \emph{Hausdorff Distance Minimization}, we regard the predicted events as a set and the ground-truth events as another set, and compute the Hausdorff distance between these two sets, which simultaneously consider all events and all their arguments. We then minimize the Hausdorff distance via gradient descent, where the model is trained to directly produce a globally optimal solution without the need of using decoding strategies as in existing approaches. 

In this way, our model learns globally and does not suffer from the problem of existing approaches that decode events based on local entity information. 
Each entity is linked to every proxy node, and the association between an entity and a proxy node is updated at each training iteration. As such, our model avoids the error propagation problem caused by the iterative decoding strategy. In addition, our approach naturally addresses the problem that the same entity mention may be involved in multiple events since the entity will be mapped to a different event-level metric space depending on its associated proxy node. 
Moreover, as our approach replaces iterative computation in decoding with parallel computation, it is computationally more efficient compared to existing path-expansion approaches, as will be shown in our experiments section. In summary, our main contributions are:
\begin{itemize}[noitemsep]
\item 
We propose a new framework for document-level multi-event extraction by learning 
event proxy nodes
in a new event-level metric space to better model the interactions among events.
\item 
We propose to utilize the Hausdorff distance in our learning objective function to optimize the difference between the generated events and the gold standard events directly.
The proposed mechanism not only simultaneously considers all events
but also speeds up the training process.
\item 
Experimental results show that our model outperforms previous state-of-the-art method 
in F1 on two datasets with only a fraction of training time.
\end{itemize}

\begin{figure*}[t]
	\centering 
	\centerline{\includegraphics[width=0.99\textwidth]{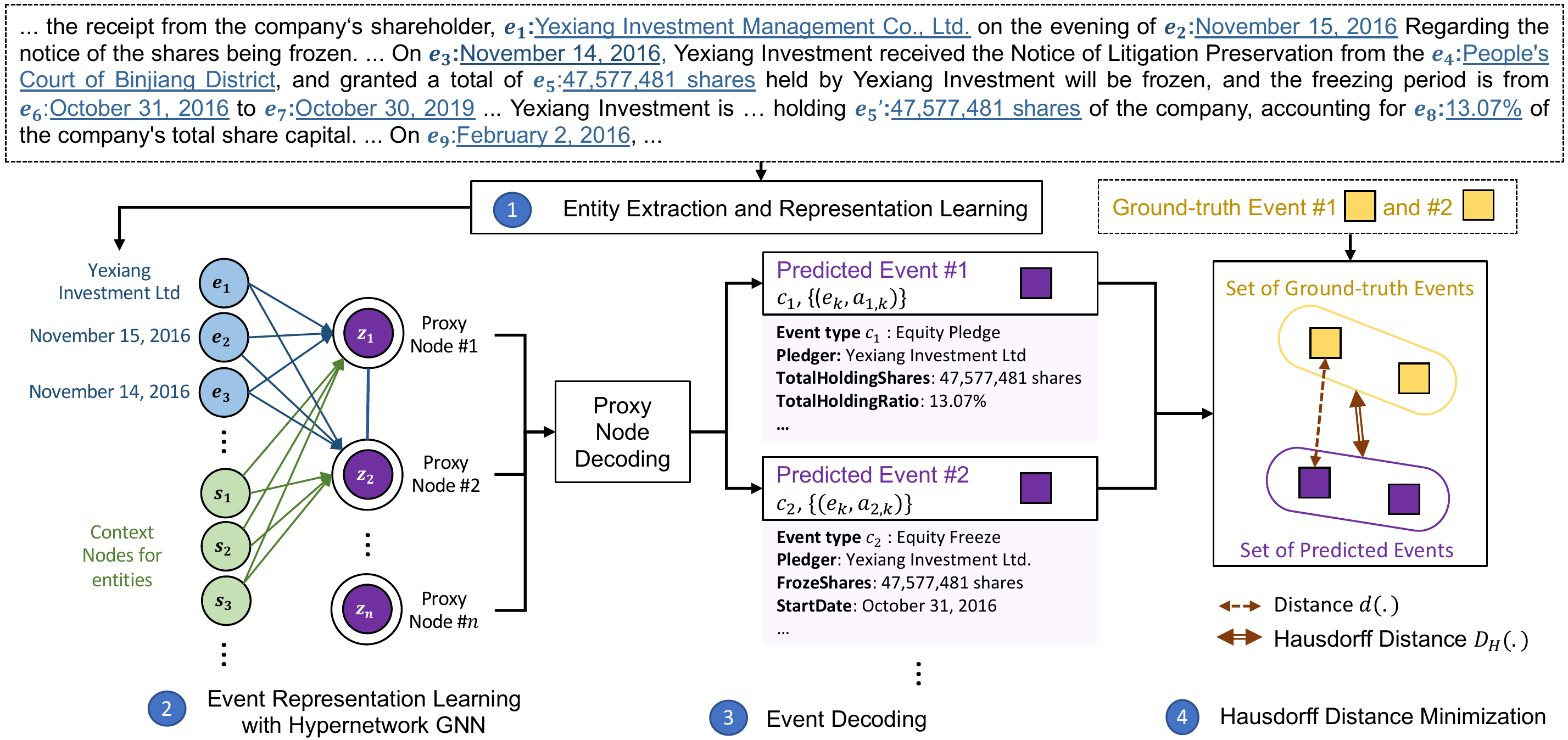}}
	\caption{Overview of \ModelName\ with the example in Figure~\ref{fig:example}, where Entity 1, 5, 8, 9 are arguments of Event~\#1; Entity 1, 4, 5, 6, 7, 8 are arguments of Event~\#2; Entity 2, 3 do not belong to any events. Before training, proxy node embeddings are randomly initialized. Entities are first mapped to entity representations in the entity-level space by \emph{Entity Representation Learning}. Then in \emph{Event Representation Learning}, a hypernetwork heterogeneous graph is constructed with entity and context nodes connected with proxy nodes, and proxy nodes interconnected with each other. Proxy nodes are updated to represent pseudo-events. Afterwards, the proxy nodes and entity nodes are decoded into events, each of which is represented by an event types and a set of argument role-entity pairs in the \emph{Event Decoding} step. Finally, \emph{Hausdorff Distance Minimization} minimizes the distance between the set of predicted events and the set of ground-truth events to perform a global training in the new event-level metric space.}
	\label{fig:overview}
\end{figure*}

\section{Related Work}

Early research on event extraction (EE) largely focused on sentence-level event extraction (SEE), aiming to classify the event trigger and arguments in a sentence. \citet{chen-etal-2015-event} decomposes SEE into two sub-tasks: \emph{event trigger detection} and \emph{event argument labeling}. More work has been done on joint-learning of the two sub-tasks \citep{
Nguyen2019OneFA, lin-etal-2020-joint}. Recently, multi-turn Question-Answer (QA) methods have been investigated for EE 
with hand-designed or automatically generated questions \citep{du-cardie-2020-event, li-etal-2020-event, wang-etal-2020-biomedical, liu-etal-2020-event, lyu-etal-2021-zero}. 
Apart from QA-based approaches, sequence-to-sequence learning has also been explored, where the event annotation is flattened as a sequence \citep{tanl_s2s_2021, lu-etal-2021-text2event, li-etal-2021-document, lu-etal-2022-unified}. More recently, prompt-based learning has been explored using the knowledge in pre-trained language models 
\citep{lin2021eliciting, hsu2021degree, ma-etal-2022-prompt}.

Compared to SEE, document-level event extraction (DEE) appears to be more challenging. 
DEE requires methods to model long-term dependencies among entities across multiple sentences. Simply employing SEE approaches for DEE may lead to incomplete and uninformative extractions \citep{li-etal-2021-document}. 
To address the problem, conditional generation have been proposed, which are conditioned on pre-specified templates or prompts
\citep{
du-etal-2021-grit, huang-etal-2021-document, ma-etal-2022-prompt}.

DEE can also be formulated as a table-filling task where each event is represented as a cluster of arguments and an event type. In such a setup, it is usually not possible to associate a particular event trigger word or phrase with an event. 
\citet{yang-etal-2018-dcfee} proposed a key-event detection model. 
\citet{zheng-etal-2019-doc2edag} transformed event tables into a directed acyclic graph with path expansion.
\citet{huang-jia-2021-exploring-sentence} constructed a graph to build sentence communities. 
\citet{Lu2022ExplainableDE} captured event clues as a series of intermediate results. 
\citet{xu-etal-2021-document} constructed a heterogeneous GNN with a tracker mechanism for partially decoded events. 
\citet{liang-etal-2022-raat} modeled the relation between entities with Relation-augmented Attention Transformer.
These methods mainly focus on modeling entity inter-relations and rely on carefully-designed event decoding strategies. 
In contrast, we model events in the event-level metric space within a more global view and with less training time.

\section{Methodology}

\subsection{Problem Setup}

Different from the trigger-based event extraction task, where an event is represented by a trigger and a list of arguments, in our task, an event is defined by an event type category $c$, a list of entities $\{e_i\}$ and their corresponding argument types $\{a_i\}$ as shown in Figure \ref{fig:example}. 
Therefore, the target output is a list of ``entity-argument'' pairs $\{(e_i, a_i)\}$ and $c$ as $\big(c,\{(e_i, a_i)\}\big)$.
Proxy node is defined as $z$. 
An overview of \ModelName\ is shown in Figure~\ref{fig:overview}. In what follows, we present each module in detail.

\subsection{Entity Representation Learning}

Given an input document, the first step is to identify the entities which might be potential arguments. 
This can be framed as a sequence labeling problem that, given a word sequence, the entity recognition model outputs a label sequence with the BIO (Beginning and Inside of an entity span, and Other tokens) tagging. We use BERT \citep{devlin-etal-2019-bert} as a sequence labeler to detect entities at sentence-level. As an entity span may contain multiple tokens, we drive its representation by averaging the hidden states of its constituent tokens. 
For a document, a total of $|e|$ entity representations are extracted as $\{\bm{h}_{e_i}\}_{i=1}^{|e|}$. The loss of the BIO sequence tagging is defined as $\mathcal{L}_{\text{er}}$. 

In order to make the entity representations encode the knowledge of entity associations, we introduce a simple auxiliary learning task to predict whether two entities belong to the same event, where entity representations will be updated during learning. 
Specifically, it is a binary classification task, with the predicted output computed as:
\begin{equation} 
	\hat{y}_{\text{epc}_{(i,j)}} = \phi \left ( \text{MLP} ( [\bm{h}_{e_i} ; \bm{h}_{e_j}] ) \right ),
\end{equation}
where $\phi$ denotes the sigmoid function, $[;]$ denotes the concatenation, and $\hat{y}_{\text{epc}_{i,j}}$ indicates the probability if entities $i$ and $j$ are from the same event. 
We use the binary cross-entropy (CE) loss here:
\begin{equation} 
    \begin{split}
    	\mathcal{L}_{\text{epc}} =  - \sum_i \sum_j \text{CE}(y_{\text{epc}_{(i,j)}}, \hat{y}_{\text{epc}_{(i,j)}})
	\end{split}
\end{equation}
where 
$y_{\text{epc}_{i,j}}$ is the label.
The loss for entity representation learning is defined as $\mathcal{L}_{\text{e}} = \mathcal{L}_{\text{er}} + \mathcal{L}_{\text{epc}}$.

\subsection{Event Representation Learning with Proxy Nodes}

In this section, we construct a graph to map entity representations in the entity-level space into event representations in a new event-level metric space.

We define $n$ proxy nodes, which serve as pseudo-events, and randomly initialize their embeddings $\{\bm{h}_{z_i}^{(0)}\}_{i=1}^{n}$,
which are only initialized once before training and will be updated during training. 
$n$ is a hyper-parameter and can be simply set to a much larger value than the expected number of extracted events, as proxy nodes can also represent \emph{null} events (see Section~\ref{sec:event_decoding}). 
We initialize entity node embeddings $\{\bm{h}_{e_i}\}_{i=1}^{|e|}$ and context node embeddings $\{\bm{h}_{s_i}\}_{i=1}^{|s|}$ by their corresponding entity and \texttt{[CLS]} representations, respectively. 

We define the graph as $\mathcal{G}=(\mathcal{V}, \mathcal{E})$, and the node set $\mathcal{V}$ contains proxy nodes, entity nodes, and context nodes as: $\mathcal{V} = \{{z_i}\}_{i=1}^{n} \cup \{{e_i}\}_{i=1}^{|e|} \cup \{{s_i}\}_{i=1}^{|s|}$ with their embeddings $\{\bm{h}_{z_i}^{(0)}\}_{i=1}^{n} \cup \{\bm{h}_{e_i}\}_{i=1}^{|e|} \cup \{\bm{h}_{s_i}\}_{i=1}^{|s|}$. The edge set $\mathcal{E}$ includes three kinds of edges as follows:


\paragraph{Proxy$\leftrightarrow$Proxy Edge} The bidirectional edge between all proxy nodes $\{z_i \rightarrow z_j: 0 < i \leq n, 0 < j \leq n\}$ allows the information exchange between proxy nodes.

\paragraph{Entity$\rightarrow$Proxy Edge} The directed edge from all entity nodes $e$ to all proxy nodes $z$ as $\{e_j \rightarrow z_i: 0 < i \leq n, 0 < j \leq |e|\}$ provides the entity information for pseudo-events.

\paragraph{Context$\rightarrow$Proxy Edge} The directed edge from all context node $s$ to all proxy node $z$ as $\{s_j \rightarrow z_i: 0 < i \leq n, 0 < j \leq |s|\}$ provides the contextual information.

In a typical setup for GNN, each node has its embedding updated by aggregating the neighborhood information. The aggregation weight matrix is shared across all nodes. But in our task here, each proxy node is expected to represent a distinct event. As such, we would like to have a unique aggregation function for each proxy node. 
To this end, we use the Graph Neural Network with Feature-wise Linear Modulation (GNN-FiLM) \citep{Brockschmidt2020GNNFiLMGN} to update the proxy node embeddings in $\mathcal{G}$. It introduces Hypernetwork to enable each proxy node to compute a unique aggregation function with different parameters. 
More concretely, given a node $v \in \mathcal{V}$ at the $(l+1)$-th layer, its hidden representation $\bm{h}_v^{(l+1)}$ is updated as:
\begin{equation} 
    \begin{aligned}
    	& \bm{h}_v^{(l+1)} = \sigma \bigg( \sum_{u \stackrel{\varepsilon}{\longrightarrow} v} \bm{\gamma}_{\varepsilon, v}^{(l)} \odot \bm{W}_{\varepsilon} \bm{h}_{u}^{(l)} + \bm{\beta}_{\varepsilon, v}^{(l)} \bigg), \\
    	& \bm{\gamma}_{\varepsilon, v}^{(l)} = f_{\gamma}(\bm{h}_{v}^{(l)}; \bm{\theta}_{\gamma, \varepsilon}), \quad
    	 \bm{\beta}_{\varepsilon, v}^{(l)} = f_{\beta}(\bm{h}_{v}^{(l)}; \bm{\theta}_{\beta, \varepsilon}),
    \end{aligned}
    \label{eq:film-general}
\end{equation}
where $u \stackrel{\varepsilon}{\longrightarrow} v$ denotes a neighboring node $u$ connected with node $v$ with the edge type $\varepsilon$. $\bm{W}_{\varepsilon} \in \mathbb{R}^{d_h \times d_h}$ is a learnable parameter for edge type $\varepsilon$. $\sigma$ and $\odot$ denote the activation function and Hadamard product, respectively. $\bm{\gamma}_{\varepsilon, v}^{(l)}$ and $\bm{\beta}_{\varepsilon, v}^{(l)}$ define the message-passing function of edge type $\varepsilon$ and node $v$ at layer $l$. They are computed by functions $f_{\gamma}$ and $f_{\beta}$ given $\bm{h}_{v}^{(l)}$ as the input. $\bm{\theta}_{\gamma, \varepsilon}$ and $\bm{\theta}_{\beta, \varepsilon}$ are learnable parameters of $f_{\gamma}$ and $f_{\beta}$,
respectively. To keep it simple, we only use one-layer GNN-FiLM with a single linear layer as the hyper-function in our experiments.

With the above formulation, each proxy node $z$ has its unique message-passing function to aggregate information from entity nodes and context nodes in different ways. 
In summary, the representations of proxy nodes $\{\widehat{\bm{h}}_{z_i}\}_{i=1}^{n}$ are updated through GNN-FiLM learning:
\begin{equation} 
	\{\widehat{\bm{h}}_{z_i}\}_{i=1}^{n} = \text{GNN-FiLM}(\mathcal{V},\mathcal{E})
\end{equation}
where $z_i$ represents a pseudo-event.
The training with proxy nodes is challenging, which will be addressed in Section~\ref{sec:minimization}.

\subsection{Event Decoding}
\label{sec:event_decoding}

In this section, each proxy node representation $\widehat{\bm{h}}_{z_i}$ is decoded into an event, which is formulated into two parallel sub-tasks: \emph{event type classification} and \emph{event argument classification}. 

\paragraph{Event Type Classification} The event type of proxy node $z_i$ is inferred from $\widehat{\bm{h}}_{z_i}$ with MLP as:
\begin{equation} 
	\bm{p}_{\hat{c}_i} = \text{softmax} \left ( \text{MLP} ( \widehat{\bm{h}}_{z_i} ) \right ), 
	\label{eq:predicted_et}
\end{equation}
where $\bm{p}_{\hat{c}_i}$ denotes the event type probability distribution of $z_i$. Event type labels includes a \emph{null} event type, denoting no correspondence between a proxy node and any events. The number of non-\emph{null} proxy nodes is the number of predicted events. 

\paragraph{Event Argument Classification} In this task, we need to associate an entity with an event under an event-specific argument type. As the same entity (e.g., a company name) may have multiple mentions 
in a document, we aggregate their representations by a Multi-Head Attention (MHA) mechanism using a proxy node as the query. 
More concretely, assuming $\{\bm{h}_{e}\}_{e \in \bar{e}_k}$ denotes a set of mentions representations for the same entity $\bar{e}_k$, we use MHA to derive the aggregated entity representation for $\bar{e}_k$. The query, key and value are defined as $\bm{Q}_{z_i} = \widehat{\bm{h}}_{z_i}, \bm{K}_{\bar{e}_k} = \{\bm{h}_{e}\}_{e \in \bar{e}_k}, \bm{V}_{\bar{e}_k} = \{\bm{h}_{e}\}_{e \in \bar{e}_k}$.
The representation of $\bar{e}_k$ is:
\begin{equation} 
	\widehat{\bm{h}}_{z_i,\bar{e}_k} = \text{MHA}(\bm{Q}_{z_i}, \bm{K}_{\bar{e}_k}, \bm{V}_{\bar{e}_k}),
\end{equation}
where $\widehat{\bm{h}}_{z_i,\bar{e}_k}$ denotes the aggregated representation for entity $\bar{e}_k$ using the proxy node $z_i$ as the query.
Then the probability distribution $\bm{p}_{\hat{a}_{i,k}}$ of argument types of entity $\bar{e}_k$ with respect to proxy node $z_i$ is:
\begin{equation} 
	\bm{p}_{\hat{a}_{i,k}} = \text{softmax} \left ( \text{MLP} ( [\widehat{\bm{h}}_{z_i} ; \widehat{\bm{h}}_{z_i,\bar{e}_k}] ) \right ),
	\label{eq:predicted_pat}
\end{equation}
where $[;]$ denotes the concatenation. The argument type set includes a \emph{null} argument type, denoting that entity $\bar{e}_k$ does not relate to proxy node $z_i$.

The final event type $\hat{c}_i$ for proxy node $z_i$ and argument type for entity $\bar{e}_k$ under the event encoded by proxy node $z_i$ are determined by:
\begin{equation} 
\begin{split}
  \hat{c}_{i} & = \text{argmax}(\bm{p}_{\hat{c}_{i}}) \\
   \hat{a}_{i,k} & = \text{argmax}(\bm{p}_{\hat{a}_{i,k}})
   \end{split}
   \label{eq:predicted_y}
\end{equation}
Each event is represented by an event type $\hat{c}_{i}$ and a list of arguments $\{\hat{a}_{i,k}\}$.
Any predicted argument type which is not in the pre-defined schema for its associated event type will be removed. Proxy nodes classified as \emph{null} event or entities classified as \emph{null} arguments will be removed. If there are multiple entities predicted as the same argument, the one with the highest probability will be kept. 

\subsection{Hausdorff Distance Minimization}
\label{sec:minimization}

In this section, 
we construct a predicted pseudo-event set $\mathcal{U}_{z}$ represented by proxy node and a ground-truth event set $\mathcal{U}_{y}$.
We define $\mu_{z_i}$ as the $i$-th pseudo-event, represented by $z_i$, with $\big(\hat{c}_i,\{(e_k, \hat{a}_{i,k})\}\big)$, and $\mu_{y_i}$ denotes the $j$-th ground-truth event $\big(c_j,\{(e_k, a_{j,k})\}\big)$.
We further define the distance $d(\mu_{z_i}, \mu_{y_j})$ between predicted event $\mu_{z_i}$ and the ground-truth event $\mu_{y_i}$ as:
\begin{equation} 
\begin{split}
    d(\mu_{z_i}, \mu_{y_j}) & = \text{CE}(\bm{p}_{\hat{c}_i}, c_j) \\ & + \frac{1}{|\bar{e}|}\sum_{k=1}^{|e|} \text{CE}(\bm{p}_{\hat{a}_{i,k}}, a_{j,k})
\end{split}
\label{eq:distance_loss}
\end{equation}
where $\text{CE}(.)$ is the cross-entropy loss; $|\bar{e}|$ denotes the number of unique entities; $k$ indicates different entities. $d(\mu_{z}, \mu_{y})$ is essentially computed by the total cross-entropy loss of event type classification and argument classification between the $i$-th proxy node and the $j$-th ground-truth event.

We aim to minimize the Hausdorff distance between sets $\mathcal{U}_{z}$ and $\mathcal{U}_{y}$ to learn the model by considering all events and their arguments simultaneously. 
As the standard Hausdorff distance is highly sensitive to outliers, we use the average Hausdorff distance \citep{Schtze2012UsingTA, Taha2015MetricsFE}:
\begin{equation} 
\begin{split}
	D_{H}(\mathcal{U}_{z}, \mathcal{U}_{y}) = \frac{1}{|\mathcal{U}_{z}|} \sum\limits_{\mu_{z} \in \mathcal{U}_{z}} \mathop{\text{min}}\limits_{\mu_{y} \in \mathcal{U}_{y}} d(\mu_{z}, \mu_{y}) \\ + \frac{1}{|\mathcal{U}_{y}|} \sum\limits_{\mu_{y} \in \mathcal{U}_{y}} \mathop{\text{min}}\limits_{\mu_{z} \in \mathcal{U}_{z}} d(\mu_{z}, \mu_{y})
	\label{eq:Hausdorff-distance2}
\end{split}
\end{equation}
However, in our task, the average Hausdorff distance could suffer a problem that a predicted event, represented by a proxy node, may be guided to learn towards more than one different t event at the same training iteration when this proxy node is the closest neighbor of multiple ground-truth events. 

To address this problem, we add a constraint to the average Hausdorff distance that the distance computation of $d(.)$ should only be performed no more than once on each $\mu_{z}$ and $\mu_{y}$, and we modify the average Hausdorff distance as:
\begin{equation} 
	\widehat{D}_{H}(\mathcal{U}_{z}, \mathcal{U}_{y}) = \text{min} \left \{ \sum\limits_{(\mu_{z},\mu_{p}) \in \mathcal{U}_{z} \times \mathcal{U}_{y}} d(\mu_{z}, \mu_{y}) \right \}
	\label{eq:Hausdorff-distance3}
\end{equation}
For example, if $d(\mu_{z_1}, \mu_{y_1})$ has been computed, then $d(\mu_{z_2}, \mu_{y_1})$ is no longer allowed to perform, as $\mu_{y_1}$ has been used in $d(.)$ computation.

To this end, Eq.~(\ref{eq:Hausdorff-distance3}) with the constraint becomes a minimum loss alignment problem. To better solve Eq.~(\ref{eq:Hausdorff-distance3}) under the constraint, we construct an undirected bipartite graph $G = (\mathcal{U}_{z}, \mathcal{U}_{y}, \mathcal{T})$, where $ \mu_{z} \in \mathcal{U}_{z}$ and $\mu_{y} \in \mathcal{U}_{y}$ are nodes of two parts representing the predicted events and the ground-truth events, respectively. $t \in \mathcal{T}$ denotes edge, which only exists between $\mu_z$ and $\mu_y$. The weight of edge $t$ between nodes $\mu_z$ and $\mu_y$ is defined as:
\begin{equation} 
    w(t_{z,y}) = d(\mu_{z}, \mu_{y})
    \label{eq:define-t}
\end{equation}
The first step is to find an edge set $\mathcal{T}$ that achieves the minimum value in the following equation:
\begin{equation} 
	\widehat{\mathcal{T}} = \text{argmin} \sum_{t_{z,y} \in \mathcal{T}} w(t_{z,y}),
	\label{eq:find-lowest-T}
\end{equation}
where the edge $t \in \mathcal{T}$ must meet these conditions:
(1) each $\mu_z$ has exactly one edge connected to it; (2) each $\mu_y$ has no more than one edge connected to it. 
Eq.~(\ref{eq:find-lowest-T}) can be computed efficiently with \citep{Ramakrishnan1991AnAD, Bertsekas1981ANA}. 
Then the final distance is computed by combining Eq.~(\ref{eq:Hausdorff-distance3}), (\ref{eq:define-t}), and (\ref{eq:find-lowest-T}) as:
\begin{equation} 
	\widehat{D}_{H}(\mathcal{U}_{z}, \mathcal{U}_{y}) = \sum_{t_{z,y} \in \widehat{\mathcal{T}}} w(t_{z,y})
\end{equation}
Finally, we use $\widehat{D}_{H}(\mathcal{U}_{z}, \mathcal{U}_{y})$ to approximate average Hausdorff distance $D_{H}(\mathcal{U}_{z}, \mathcal{U}_{y})$.

As $n$ has been set to be a very large number, if the number of ground-truth events is less than the number of predicted events in a document, pseudo \emph{null} events are added to the ground-truth event set as negative labels to make the number of ground-truth events equals to the number of predicted events.

In summary, $\widehat{D}_{H}(\mathcal{U}_{z}, \mathcal{U}_{y})$ is the distance between the predicted events set and the ground-truth events set, which considers all events with all of their arguments at the same time, essentially capturing a global alignment. 

\subsection{Objective Function}

The final loss is the sum of approximate Hausdorff distance and entity representation loss:
\begin{equation} 
	\mathcal{L} = \widehat{D}_{H}(\mathcal{U}_{z}, \mathcal{U}_{y}) + 
 \mathcal{L}_{\text{e}}
	\label{eq:final-loss}
\end{equation}

\section{Experiments}

In this section, we present performance and run-time experiments in comparison with state-of-the-art approaches. 
We also discuss the ablations study. Entity and event visualisation results can be found in Appendix~\ref{app:visualisation}.

\begin{table*}[th]
\centering
\resizebox{0.9\linewidth}{!}{
\begin{tabular}{@{}lcccccccccc@{}}
\toprule
\multirow{2}{*}{\textbf{Model}}  & \multicolumn{5}{c}{\textbf{ChFinAnn}} & \multicolumn{5}{c}{\textbf{DuEE-Fin}} \\
\cmidrule(lr){2-6} \cmidrule(lr){7-11} 
& \textbf{P.}   & \textbf{R.}   & \textbf{F1}   & \textbf{F1 (S.)}   & \textbf{F1 (M.)} &\textbf{P.}   & \textbf{R.}   & \textbf{F1}   & \textbf{F1 (S.)}   & \textbf{F1 (M.)} \\ 
\midrule
DCFEE-O & 68.0 & 63.3 & 65.6 & 69.9 & 50.3 &59.8 & 55.5 & 57.6 & 62.7 & 53.3\\
DCFEE-M & 63.0 & 64.6 & 63.8 & 65.5 & 50.5 & 50.2 & 55.5 & 52.7 & 57.1 & 49.5 \\
Greedy-Dec & 82.5 & 53.7 & 65.1 & 80.2 & 36.9  & 66.0 & 50.6 & 57.3 & 67.8 & 47.4 \\
Doc2EDAG & 82.7 & 75.2 & 78.8 & 83.9 & 67.3 & 67.1 & 60.1 & 63.4 & 69.1 & 58.7 \\
DE-PPN & 83.7 & 76.4 & 79.9 & 85.9 & 68.4 & 69.0 & 33.5 & 45.1 & 54.2 & 21.8 \\
PTPCG & 83.7 & 75.4 & 79.4 & 88.2 & - & 71.0 & 61.7 & 66.0 & - & - \\
GIT & 82.3 & 78.4 & 80.3 & 87.6 & 72.3 & 69.8 & 65.9 & 67.8 & 73.7 & 63.8 \\
ReDEE & 83.9 & 79.9 & 81.9 & 88.7 & 74.1 & 77.0 & 72.0 & 74.4 & 78.9 & 70.6 \\
\midrule
\ModelName\ (Ours) & \textbf{84.1} & \textbf{81.9} & \textbf{83.0} & \textbf{89.6} & \textbf{75.6} & \textbf{78.8} & \textbf{72.8} & \textbf{75.6} & \textbf{80.0} & \textbf{72.1} \\
\bottomrule
\end{tabular} 
}
\caption{\label{tab:doc2edag_du_overall} Overall precision (P.), recall (R.), and F1-score (F1) on the ChFinAnn and DuEE-Fin datasets. F1 (S.) and F1 (M.) denote scores under Single-event (S.) and Multi-event (M.) sets.}
\end{table*}

\subsection{Experimental Setup}



\paragraph{Dataset} We evaluate \ModelName\ on the two document-level multi-event extraction datasets: 
\underline{(1) ChFinAnn dataset}\footnote{\url{https://github.com/dolphin-zs/Doc2EDAG}} \citep{zheng-etal-2019-doc2edag} 
consists of  32,040 financial documents, with 25,632, 3,204, and 3,204 in the train, development, and test sets, respectively, 
and includes five event types.
The dataset contains 71\% of single-event documents and 29\% of multi-event documents.
\underline{(2) DuEE-Fin dataset}\footnote{\url{https://aistudio.baidu.com/aistudio/competition/detail/46/0/task-definition}} \citep{10.1007/978-3-031-17120-8_14} 
has around 11,900 financial documents and 13 event types.
\wxytodo{As the released dataset does not contain the ground truth annotations for the test set, we follow the setting of \cite{liang-etal-2022-raat} and use the original development set as the test set. We set aside 500 documents from the training set as the development set. Our final dataset has 6,515, 500, and 1,171 documents in the train, development, and test sets, respectively.}
There are 67\% of single-event documents and 33\% of multi-event documents.
More details about the event types and their distributions are in Appendix~\ref{app:dataset}.

\paragraph{Evaluation Metrics} We follow the same metrics in \citep{zheng-etal-2019-doc2edag}. For a predicted event of a specific event type, the most similar ground-truth event that is of the same event type is selected without replacement. Then the micro-averaged role-level precision, recall, and F1-score are calculated for the predicted event and the selected gold event.

\paragraph{Implementation Detail} 
To keep it simple, we only use one-layer GNN-FiLM \citep{Brockschmidt2020GNNFiLMGN} with a single linear layer as the hyper-function.
Specifically, we have
$f_{\gamma}(\bm{h}_{v}^{(l)}; \bm{\theta}_{\gamma, \varepsilon}) = \bm{W}_{\gamma, \varepsilon} \bm{h}_{v}^{(l)}$ and $f_{\beta}(\bm{h}_{v}^{(l)}; \bm{\theta}_{\beta, \varepsilon}) = \bm{W}_{\beta, \varepsilon} \bm{h}_{v}^{(l)}$
in Eq.~(\ref{eq:film-general}). 
The number of proxy nodes $n$ is set to 16. 
More implementation details are in Appendix~\ref{app:implementation}


\paragraph{Baselines}

The baselines that we compare with are as follows: 
\textbf{DCFEE} \citep{yang-etal-2018-dcfee} uses an argument-completion strategy in the table-filling task. Two variants of DCFEE are \textbf{DCFEE-O} for single-event and \textbf{DCFEE-M} for multi-event. \textbf{Doc2EDAG} \citep{zheng-etal-2019-doc2edag} utilizes a path-expansion decoding strategy to extract events like hierarchical clustering. \textbf{Greedy-Dec} is a variant of Doc2EDAG that decodes events greedily. \textbf{DE-PPN} \citep{yang-etal-2021-document} uses Transformer to encode sentences and entities. 
\textbf{GIT} \citep{xu-etal-2021-document} 
uses a Tracker module to track events in the path-expansion decoding. 
\textbf{PTPCG} \citep{ijcai2022p0632} combines event arguments together in a non-autoregressive decoding approach with pruned complete graphs, aiming to consume lower computational resources. 
\textbf{ReDEE} \citep{liang-etal-2022-raat} is a Relation-augmented Attention Transformer to cover multi-scale and multi-amount relations.

\begin{table}[h]
\centering
\resizebox{0.9\columnwidth}{!}{
\begin{tabular}{@{}lccccc@{}}
\toprule
\textbf{Model} & \textbf{EF}   & \textbf{ER}   & \textbf{EU}   & \textbf{EO}   & \textbf{EP} \\ 
\midrule
DCFEE-O & 51.1 & 83.1 & 45.3 & 46.6 & 63.9\\
DCFEE-M & 45.6 & 80.8 & 44.2 & 44.9 & 62.9 \\
Greedy-Dec & 58.9 & 78.9 & 51.2 & 51.3 & 62.1 \\
Doc2EDAG & 70.2 & 87.3 & 71.8 & 75.0 & 77.3 \\
DE-PPN & 73.5 & 87.4 & 74.4 & 75.8 & 78.4 \\
GIT & 73.4 & 90.8 & 74.3 & 76.3 & 77.7 \\ 
ReDEE & 74.1 & 90.7 & 75.3 & \textbf{78.1} & 80.1 \\
\midrule
\ModelName\ (Ours) & \textbf{75.7} & \textbf{93.7} & \textbf{76.0} & 72.0 & \textbf{81.3} \\ 
\bottomrule
\end{tabular} 
}
\caption{\label{tab:doc2edag_event} F1-score of five event types on ChFinAnn. 
}
\end{table}

\begin{table*}[th]
\centering
 \resizebox{0.95\linewidth}{!}{
\begin{tabular}{@{}lccccccccccccc@{}}
\toprule
\textbf{Model} & \textbf{WB} & \textbf{FL} & \textbf{BA} & \textbf{BB} & \textbf{CF} & \textbf{CL} & \textbf{SD} & \textbf{SI} & \textbf{SR} & \textbf{RT} & \textbf{PR} & \textbf{PL} & \textbf{EC} \\ 
\midrule
DCFEE-O & 54.0 & 65.4 & 44.0 & 27.3 & 58.2 & 42.0 & 48.8 & 53.9 & 76.7 & 32.9 & 63.3 & 58.3 & 40.6 \\
DCFEE-M & 49.2 & 68.0 & 40.4 & 28.4 & 51.2 & 35.1 & 42.3 & 45.9 & 74.0 & 51.0 & 55.8 & 56.4 & 37.4 \\
Greedy-Dec & 53.7 & 71.8 & 49.5 & 41.1 & 61.3 & 42.1 & 49.7 & 57.4 & 74.4 & 29.2 & 60.8 & 50.5 & 39.4 \\
Doc2EDAG & 60.0 & 78.3 & 50.6 & 40.1 & 63.2 & 51.5 & 50.7 & 52.9 & 83.7 & 51.2 & 64.8 & 61.7 & 51.2 \\
DE-PPN & 50.7 & 62.7 & 41.3 & 21.4 & 36.3 & 23.0 & 32.9 & 31.3 & 67.8 & 25.8 & 42.1 & 36.3 & 23.4 \\
GIT & 58.8 & 77.6 & 56.6 & 44.7 & 68.5 & 55.1 & 58.8 & \textbf{71.2} & 86.4 & 45.0 & 66.4 & 71.3 & 53.8 \\
ReDEE & 72.2 & 81.2 & 58.9 & 53.4 & 76.7 & 56.7 & 68.2 & 56.6 & \textbf{90.6} & 49.9 & 75.0 & \textbf{77.8} & 56.6  \\
\midrule
\ModelName\ (Ours)& \textbf{76.0} & \textbf{85.0} & \textbf{69.8} & \textbf{63.5} & \textbf{79.0} & \textbf{60.5} & \textbf{69.3} & 68.2 & 89.2 & \textbf{50.0} & \textbf{77.4} & 76.9 & \textbf{56.9} \\
\bottomrule
\end{tabular} 
}
\caption{\label{tab:duee_event} F1-score on the DuEE-Fin dataset with 13 event types.}
\end{table*}


\subsection{Overall Results}
\label{sec:experiment_results}

Table~\ref{tab:doc2edag_du_overall} shows the 
results on the ChFinAnn and the DuEE-Fin datasets. 
For ChFinAnn, the baseline results are reported in \citep{zheng-etal-2019-doc2edag, yang-etal-2021-document, xu-etal-2021-document, ijcai2022p0632, liang-etal-2022-raat}. 
For DuEE-Fin, the baseline results are either taken from \citep{liang-etal-2022-raat} or by running the published source code of the baselines. 
We can observe that a simple argument completion strategy (DCFEE-O and DCFEE-M) produces the worst results. Greedy-Dec with the greedy decoding strategy improves upon DCEFF variants, but it reached an F1-score lower than Doc2EDAG by 13.7\% on ChFinAnn and 6.3\% on DuEE-Fin due to only modeling entity-level representations without a global view. DE-PPN which uses the Transformer to encode sentences and entities performs worse compared to Doc2EDAG which utilizes a path-expansion decoding strategy. Extending DocEDAG with a Track module (GIT) or using a relation-augmented attention transformer (ReDEE) achieves better results compared to earlier approaches. 
\ModelName\ gives the best overall F1-score, outperforming the best baseline, ReDEE, by 1.1-1.2\%, respectively on ChFinAnn and DuEE-Fin. It can also be observed that all models have better F1-scores for the single-event scenario than the multi-event one, verifying the difficulty of extracting multiple events from a document. When comparing results across the two datasets, we see better results achieved on ChFinAnn, possibly due to its larger training set and smaller set of event types compared to DuEE-Fin.

\subsection{Per-Event-Type Results}
\label{sec:experiment_results_event}

Table~\ref{tab:doc2edag_event} and Table~\ref{tab:duee_event} show the evaluation results on the 5 and 13 event types\footnote{Please refer to Appendix~\ref{app:dataset} for event type descriptions.} on ChFinAnn and DuEE-Fin, respectively. 
On ChFinANN, ReDEE outperforms the others on EO. 
On DuEE-Fin, ReDEE gives the best results on SR and PL, while GIT outperforms the others on SI.
Some documents of these event types contain more than 40 sentences.
A possible reason for \ModelName\ not performing well on these event types is its limited capability of capturing long-term dependencies across sentences, since 
\ModelName\ does not directly model the relations between sentences. 
On the contrary, ReDEE and GIT model the inter-relations of sentences directly.
Nevertheless, \ModelName\ achieves superior results on other event types, resulting in overall better performance compared to baselines.

\subsection{Run-Time Comparison}



\begin{table}[h]
\centering
\resizebox{0.9\columnwidth}{!}{
\begin{tabular}{@{}lcccc@{}}
\toprule
\multirow{2}{*}{\textbf{Model}} & \multicolumn{2}{c}{\textbf{Per Epoch}} & \multicolumn{2}{c}{\textbf{Convergence}} \\
\cmidrule(lr){2-3}\cmidrule(lr){4-5}
& \textbf{Time}     & \textbf{Ratio}     & \textbf{Time}      & \textbf{Ratio}      \\ 
\midrule
\multicolumn{5}{c}{\textbf{ChFinANN}}\\
\midrule
Doc2EDAG & 4:40 & 5.2x & 327:09 & 10.7x \\
DE-PPN & 1:54 & 2.1x & 87:27 & 2.8x\\
PTPCG & 0:26 & 0.5x & 39:04 & 1.3x \\
GIT & 4:48 & 5.3x & 317:35 & 10.3x \\
ReDEE & 8:12 & 9.1x & 525:33 & 17.1x \\
\midrule
\ModelName\ (Ours) & 0:54 & 1.0x & 30:34 & 1.0x \\ 
\midrule
\multicolumn{5}{c}{\textbf{DuEE-Fin}}\\
\midrule
Doc2EDAG & 1:53 & 11.3x & 249:35 & 16.5x \\
DE-PPN & 0:15 & 1.5x & 24:36 & 1.6x\\
PTPCG & 0:06 & 0.6x & 9:28 & 0.6x \\
GIT & 1:50 & 11.0x & 178:38 & 11.8x \\
ReDEE & 7:28 & 44.8x & 687:14 & 45.4x \\
\midrule
\ModelName\ (Ours) & 0:10 & 1.0x & 15:09 & 1.0x \\ 
\bottomrule
\end{tabular} 
}
\caption{\label{tab:run-time} The GPU time (hh:mm) of each epoch and reaching convergence in average.}
\end{table}



We compare the training time of the five baselines, Doc2EDAG, DE-PPN, PTPCG, GIT, and ReDEE, with \ModelName\ 
on a GPU server with NVIDIA Quadro RTX 6000 and the same setting. 
We record the average per epoch training time 
and the total time to reach convergence in Table~\ref{tab:run-time}.
DuEE-Fin contains fewer data than ChFinANN, as such, Doc2EDAGE, GIT, and \ModelName\ trained faster on DuEE-Fin.
However, ReDEE took longer time to converge on DuEE-Fin, because ReDEE models the relations of all argument-argument pairs. As the number of event types and argument types in DuEE-Fin is more than that in ChFinANN, the training time of ReDEE increases exponentially. 
\wxyedit{
DE-PPN runs faster than Doc2EDAG, GIT, and ReDEE but slower than \ModelName .
In contrast, \ModelName\ avoids the time-consuming decoding by introducing the proxy nodes and \MinName .
Besides, \ModelName\ can process all proxy nodes and their arguments in parallel, which is more GPU-friendly.
PTPCG has a shorter per-epoch run time, but took a longer time to converge on ChFinAnn; though it appears to be more run time efficient on DuEE-Fin compared to our approach. }
In summary, \ModelName\ is 0.5x-44.8x times faster than the baselines per epoch, and is 0.6x-45.4x times faster to reach convergence.

\begin{figure*}[ht]
	\centering 
	\centerline{\includegraphics[width=0.85\textwidth]{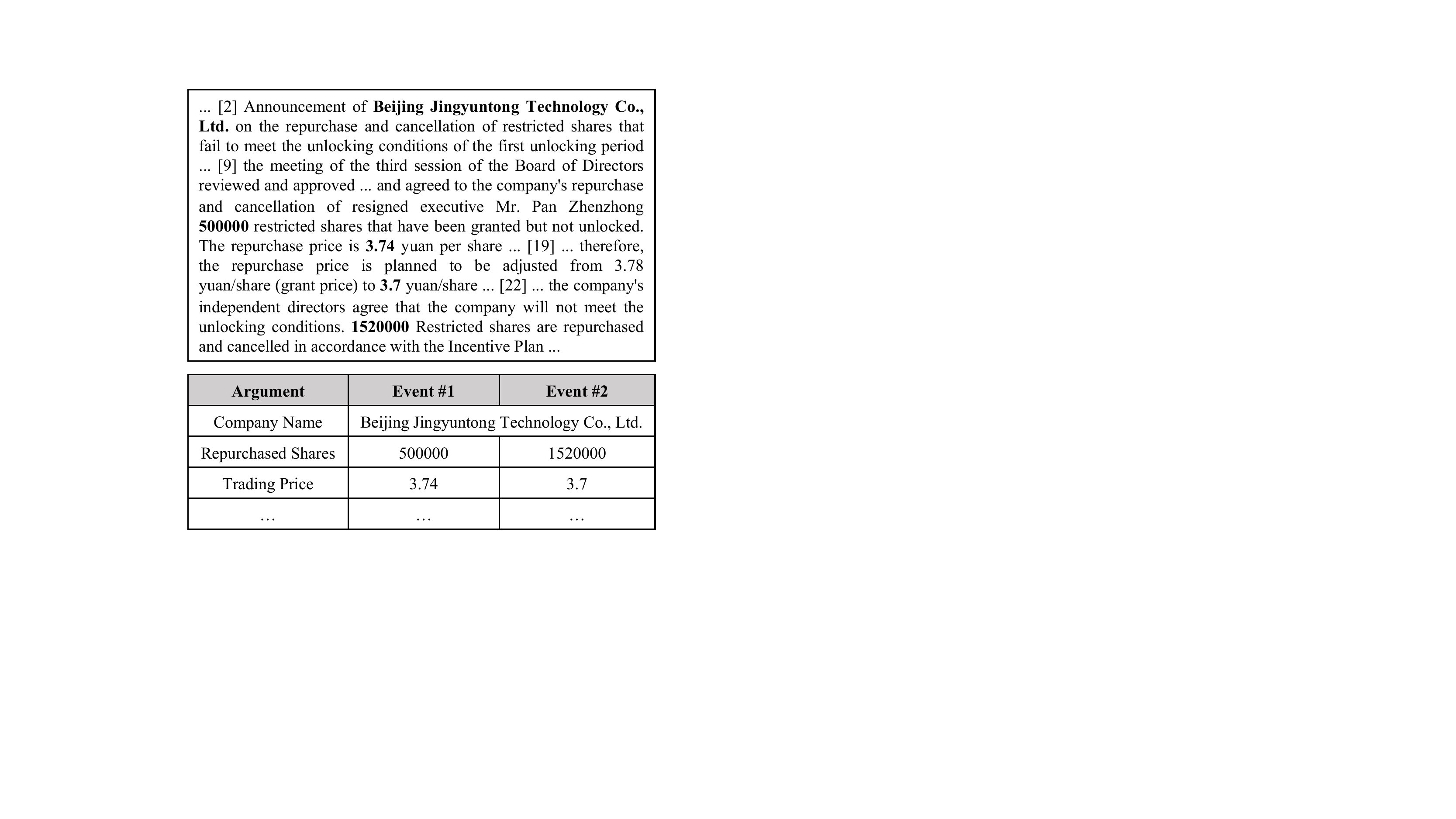}}
	\caption{Error case study with incorrect arguments colored in red. [.] denotes the sentence numbering.}
	\label{fig:case_fail}
\end{figure*}

\subsection{Ablation Study}


\begin{table}[th]
\centering
\resizebox{0.99\columnwidth}{!}{
\begin{tabular}{@{}lcccccc@{}}
\toprule
\multirow{2}{*}{\textbf{Model}} & \multicolumn{3}{c}{\textbf{ChFinANN}} & \multicolumn{3}{c}{\textbf{DuEE-Fin}} \\
\cmidrule(lr){2-4}\cmidrule(lr){5-7}
& \textbf{P.} & \textbf{R.} & \textbf{F1} & \textbf{P.} & \textbf{R.} & \textbf{F1} \\
\midrule
ProCNet (Ours) & 84.1 & 81.9 & 83.0 & 78.8 & 72.8 & 75.6 \\ 
\midrule
$-$Hypernetwork & 82.7 & 81.6 & 82.1 & 77.0 & 72.2 & 74.5 \\
$-$Proxy node & 41.3 & 2.3 & 4.4 & 21.1 & 1.0 & 1.7 \\
$-$\MinName\ & 17.0 & 19.8 & 18.3 & 13.3 & 8.2 & 10.1 \\ 
\bottomrule
\end{tabular} }
\caption{\label{tab:all_ablation} Ablation study on ChFinAnn and DuEE-Fin.}
\end{table}


Table~\ref{tab:all_ablation} shows how different components in \ModelName\ contribute to performance: 

\paragraph{$-$Hypernetwork}
Hypernetwork is removed by replacing GNN-FiLM with RGCN \citep{10.1007/978-3-319-93417-4_38_18}, 
where all proxy nodes in RGCN share the same message-passing function. We see a drop of about 1\% in F1 on both datasets, showing the importance of using different entity aggregation functions for different event proxy nodes.

\paragraph{$-$Proxy Node} 
We replace $\{\bm{h}_{z_i}\}_{i=1}^{n}$ with $\{\bm{h}_{z_0}\}_{i=1}^{n}$, where all proxy nodes share the same embedding $\bm{h}_{z_0}$. In this way, $\bm{h}_{z_0}$ acts as a common start node as in existing baselines. It can be observed that F1 drops significantly to 4.4\% and 1.7\%, respectively. The model learns almost nothing, which verifies the importance of the proxy nodes for \ModelName .

\paragraph{$-$\MinName } 
Instead of minimizing the Hausdorff distance between the predicted set and the ground-truth set globally, 
we randomly initialize the edge set $\widehat{\mathcal{T}}$ without employing Eq.~(\ref{eq:find-lowest-T}), where the minimization is not performed towards the global minimum. We see a drastic decrease in performance. 
Without \MinName , it is difficult for the the model to learn the alignment between a proxy node and a ground-truth event, showing that 
\MinName\ is an indispensable component of \ModelName .






\subsection{Case Study}
\label{app:case_study}

\wxyedit{
Figure~\ref{fig:case_fail} shows an error case of \ModelName . \emph{Event \#1} has its arguments spanned from sentence \#9 to sentence \#21. The text contains a few dates, making it difficult for our model to assign the correct dates to the \emph{StartDate} and \emph{EndDate} arguments. 
For \emph{Event \#2}, the detection of \emph{LaterHoldingShares} requires the model to associate the `\emph{above-mentioned increase}' with \emph{Event \#2}. These errors show that \ModelName still faces a difficulty in modeling long-distance dependencies. 
}

\section{Conclusion}

In this paper, we no longer focus on inter-entities relation modeling and decoding strategy 
as in previous methods, but directly learns all events globally through the use of event proxy nodes and the minimization of the Hausdorff distance in our proposed \ModelName. 
In our experiments, \ModelName\ outperforms state-of-the-art approaches while only requiring a fraction of time for training.

\section*{Acknowledgements}
This work was supported in part by the UK Engineering and Physical Sciences Research Council (grant no. EP/T017112/2, EP/V048597/1, EP/X019063/1). YH is supported by a Turing AI Fellowship funded by the UK Research and Innovation (grant no. EP/V020579/2).

\section*{Limitations}

In our proposed model, we introduce a hyper-parameter $n$ as the number of event proxy nodes. The value of $n$ needs to be pre-set. Setting $n$ to a value larger than the actual event number in a document would lead to computational redundancy as more proxy nodes would be mapped to the \emph{null} event. However, setting $n$ to a small value may miss some events in a document. We have experimented with automatically learning the value of $n$ based on an input document in \ModelName. But we did not observe improved event extraction performance. As such, we simply set it to 16. In the ChFinAnn dataset, 98\% documents have less than 7 events annotated. This results in the learning of many redundant proxy nodes for such documents. It remains an open challenge on automatically learning a varying number of event proxy nodes based on an input document. Reducing the number of redundant proxy nodes can reduce training time further.



Another shortcoming is the limited capability of \ModelName\ in capturing the long-term dependencies of sentences, 
as have been discussed in the per-event-type results in Section~\ref{sec:experiment_results}~and~\ref{sec:experiment_results_event}. 
We observed a relatively worse performance of \ModelName\ in dealing with long documents with more than 40 sentences as it does not explicitly model the inter-relations of sentences. One possible direction is to explore the use of a heterogeneous graph which additionally models the entity-entity, entity-sentence, and sentence-sentence relations. We will leave it as the future work to study the trade-off between event extraction performance and training efficiency.





\bibliography{anthology,custom}
\bibliographystyle{acl_natbib}

\clearpage
\appendix

\setcounter{figure}{0}
\renewcommand{\thefigure}{A\arabic{figure}}

\setcounter{table}{0}
\renewcommand{\thetable}{A\arabic{table}}

\section*{Appendix}

\section{Experimental Setup}
\label{app:experiment}

\subsection{Dataset}
\label{app:dataset}

\begin{table}[th]
\centering
\begin{tabular}{@{}lc@{}}
\toprule
\textbf{Event Type} & \textbf{Distribution} \\
\midrule
\textbf{E}quity \textbf{F}reeze & 4.2\% \\
\textbf{E}quity \textbf{R}epurchase & 9.5\% \\
\textbf{E}quity \textbf{U}nderweight & 16.0\% \\
\textbf{E}quity \textbf{O}verweight & 18.3\% \\
\textbf{E}quity \textbf{P}ledge & 52.0\% \\
\bottomrule
\end{tabular}
\caption{\label{tab:doc2edag_dataset} Event type distribution in ChFinAnn.}
\end{table}

\begin{table}[th]
\centering
\begin{tabular}{@{}lc@{}}
\toprule
\textbf{Event Type} & \textbf{Distribution} \\
\midrule
\textbf{W}in \textbf{B}idding & 9.5\% \\
\textbf{F}inancial \textbf{L}oss & 11.1\% \\
\textbf{B}usiness \textbf{A}cquisition & 9.7\% \\
\textbf{B}usiness \textbf{B}ankruptcy & 2.5\% \\
\textbf{C}Corporate \textbf{F}inancing & 5.5\% \\
\textbf{C}ompanies \textbf{L}isting & 5.1\% \\
\textbf{S}hareholders Holdings \textbf{D}ecrease & 9.3\% \\
\textbf{S}hareholders Holdings \textbf{I}ncrease & 3.5\% \\
\textbf{S}hare \textbf{R}epurchase & 14.1\% \\
\textbf{R}egulatory \textbf{T}alk & 1.8\% \\
\textbf{P}ledge \textbf{R}elease & 7.7\% \\
\textbf{Pl}edge & 10.8\% \\
\textbf{E}xecutive \textbf{C}hange & 9.4\% \\
\bottomrule
\end{tabular}
\caption{\label{tab:duee_dataset} Event type distribution in DuEE-Fin.}
\end{table}

\paragraph{ChFinAnn} 
ChFinAnn dataset contains 32,040 financial documents collected from public reports, with 25,632 in the train set, 3,204 in the development set and 3,204 in the test set. 
There are 71\% of single-event documents and 29\% of multi-event documents.
It includes five event types.
The distribution of event types is shown in Table~\ref{tab:doc2edag_dataset}.

\paragraph{DuEE-Fin} 
DuEE-Fin dataset did not release the ground truth publicly available for the test set. We follow the setting of \citet{liang-etal-2022-raat}, but additionally split 500 documents from train set as development set and treat the original development set as test set. To this end, there are 6,515, 500, and 1,171 documents in train, development, and test set, respectively.
There are 67\% of single-event documents and 33\% of multi-event documents.
The DuEE-Fin dataset contains 13 event types.
The distribution of event types is shown in Table~\ref{tab:duee_dataset}.

\subsection{Implementation Detail}
\label{app:implementation}

We follow the setting of \citet{liang-etal-2022-raat} using the BERT-base \citep{devlin-etal-2019-bert} in Roberta setting \citep{liu2019roberta} as the sequence labeling model. We use one-layer GNN-FiLM \citep{Brockschmidt2020GNNFiLMGN} with a single linear layer as the hyper-function and GELU \citep{Hendrycks2016GaussianEL} as the activation function. 
Specifically, we have
$f_{\gamma}(\bm{h}_{v}^{(l)}; \bm{\theta}_{\gamma, \varepsilon}) = \bm{W}_{\gamma, \varepsilon} \bm{h}_{v}^{(l)}$ and $f_{\beta}(\bm{h}_{v}^{(l)}; \bm{\theta}_{\beta, \varepsilon}) = \bm{W}_{\beta, \varepsilon} \bm{h}_{v}^{(l)}$ 
in Eq.~(\ref{eq:film-general}), where $\bm{W}_{\gamma, \varepsilon} \in \mathbb{R}^{d_h \times d_h}$ and $\bm{W}_{\beta, \varepsilon} \in \mathbb{R}^{d_h \times d_h}$ are learnable parameters. 
The hidden size is $512$. 
We employ the Adam optimizer \citep{journals/corr/KingmaB14} with a batch size 32, a learning rate \mbox{1$e$-5} for pretrained parameters, a learning rate \mbox{1$e$-4} for randomly initialized parameters. We run the model 3 times with a maximal number of epochs 100 selecting the best checkpoint and with one NVIDIA Quadro RTX 6000 GPU.

\section{Visualisation}
\label{app:visualisation}

\begin{figure*}[htb]
    \centering
    \begin{subfigure}[b]{0.4\textwidth}
        \includegraphics[width=\textwidth]{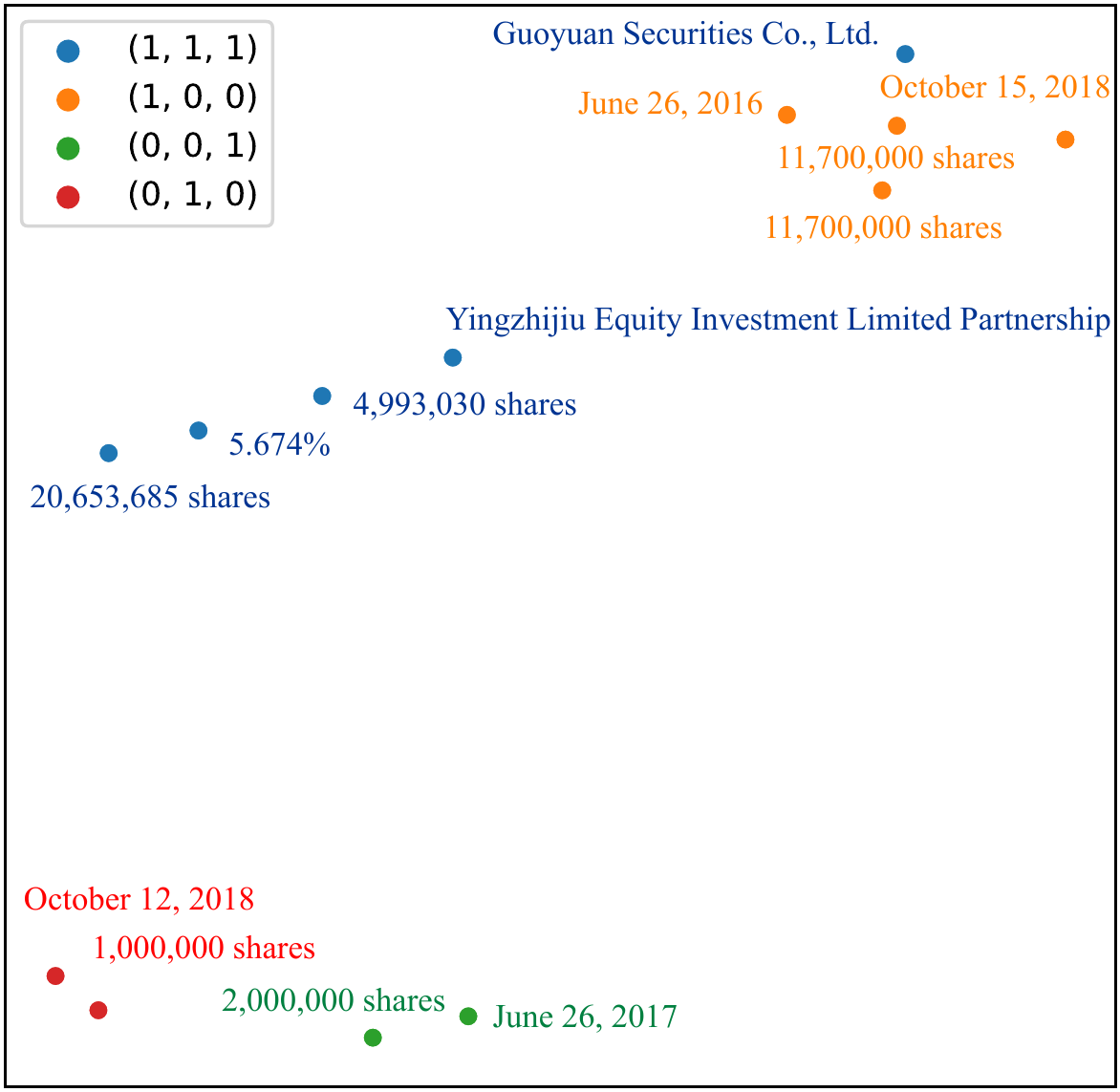}
        \caption{Entity}
        \label{fig:vis-entity}
    \end{subfigure}
    \qquad
    \begin{subfigure}[b]{0.4\textwidth}
        \includegraphics[width=\textwidth]{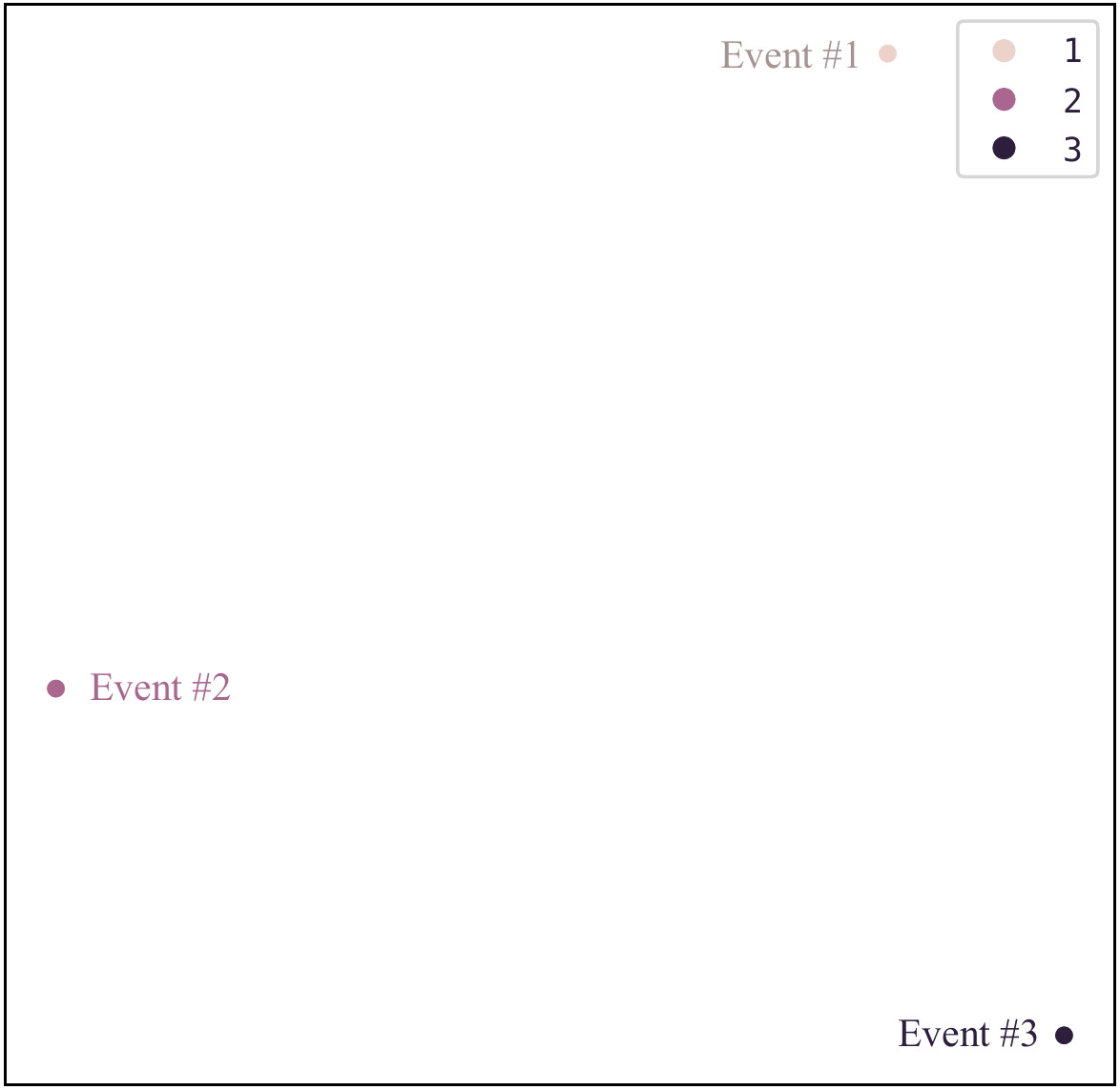}
        \caption{Proxy Node}
        \label{fig:vis-proxy-node}
    \end{subfigure}
  \caption{Visualisation of entity and proxy node representations of the example in Figure~\ref{fig:vis_example}.}
  \label{fig:visualisation}
\end{figure*}

\begin{figure*}[htb]
	\centering 
	\centerline{\includegraphics[width=0.9\textwidth]{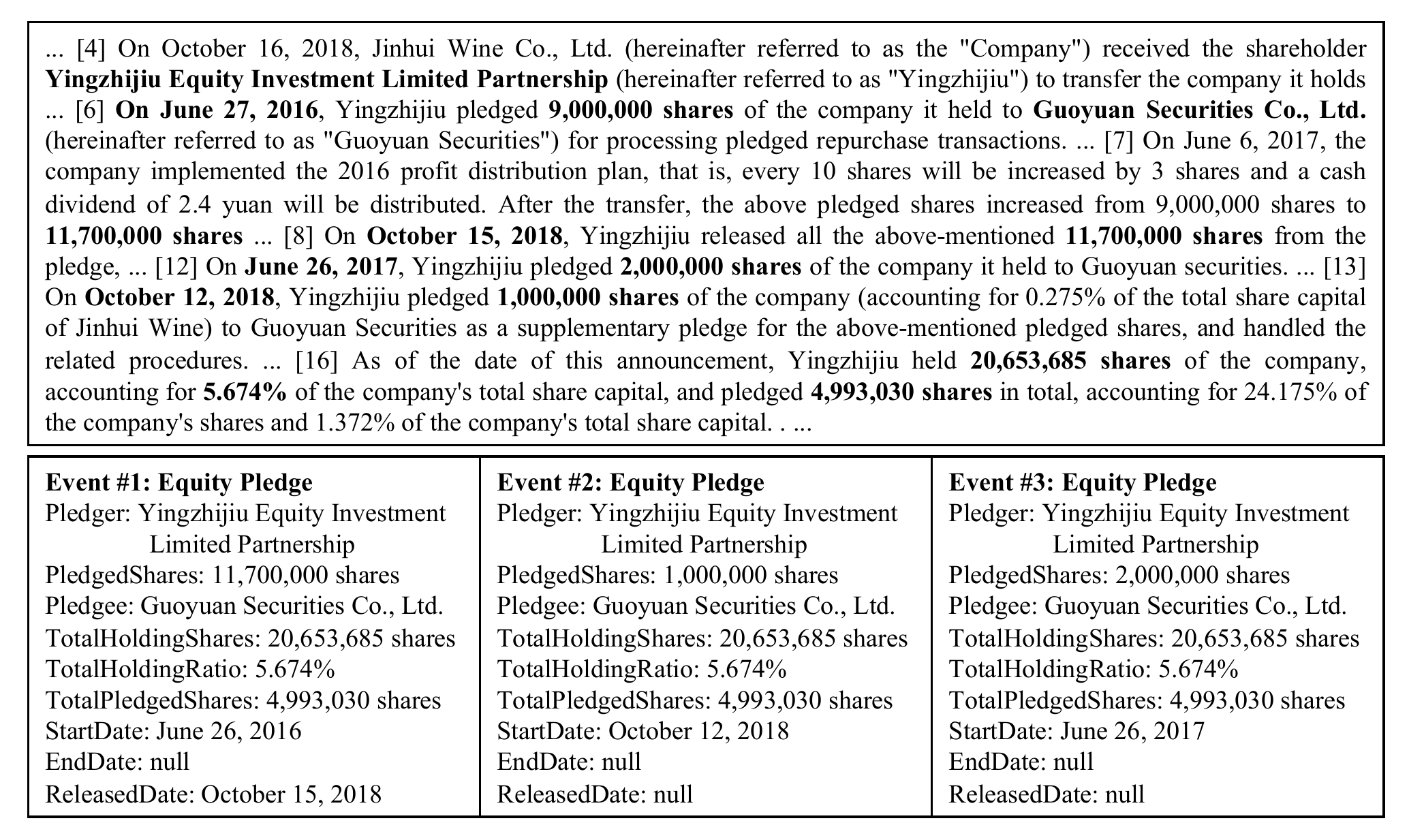}}
	\caption{The example used in Figure~\ref{fig:visualisation}. Bold words are arguments, and [.] denotes the sentence numbering.}
	\label{fig:vis_example}
\end{figure*}

We employ t-SNE \citep{Maaten2008VisualizingDU} to visualize in Figure~\ref{fig:visualisation} the representations of entities and proxy nodes of an example illustrated in Figure~\ref{fig:vis_example}. The three numbers in Figure~\ref{fig:vis-entity} denote whether an entity belongs to the three corresponding events. For example, the \emph{(0, 0, 1)} means green entities are arguments of \emph{Event \#3}, whereas \emph{(1, 1, 1)} means blue entities are arguments of all three events. Blue entities are also separated from the other three kinds of entities. It is difficult to identify events from the entity-level representations. In contrast, after mapping entities to the event-level metric space, the three points denoting the three proxy nodes are easier to be distinguished as shown in Figure~\ref{fig:vis-proxy-node}.

Figure~\ref{fig:vis_example} shows the example used in Figure~\ref{fig:visualisation}. The three events correspond to three proxy nodes. The entity \emph{11,700,000 shares} appears two times in the document, so there are two points in Figure~\ref{fig:visualisation} representing \emph{11,700,000 shares}.

\end{document}